\RequirePackage{amsmath}
\documentclass[graybox]{svmult}

\usepackage[T1]{fontenc} 

\usepackage{amssymb}

\usepackage{textcomp}
\usepackage{gensymb}

\usepackage{helvet}
\usepackage{courier}
\usepackage{type1cm}         
\usepackage{makeidx}         
\usepackage{graphicx}        
\usepackage{multicol}        
\usepackage[bottom]{footmisc}

\begin{document}

\title*{3D Convolutional Networks for Action Recognition: Application to Sport Gesture Recognition}
\titlerunning{3D CNNs for Action Recognition: Application to Sport Gesture Recognition}
\label{chapter::chapter_10}
\author{
    P.-E. Martin$^1$
    \and J. Benois-Pineau$^1$
    \and R. P\'eteri$^2$ 
    \and A. Zemmari$^1$ 
    \and J. Morlier$^3$}
    \institute{$^1$LaBRI, Univ. Bordeaux, 351 cours de la Libération, F33405 Talence Cedex, France. \\
    $^2$MIA, Univ. La Rochelle, La Rochelle, France\\
   $^3$IMS, Univ. Bordeaux, France\\
    \email{\{pierre-etienne.martin,benois-p,zemmari\}@u-bordeaux.fr, renaud.peteri@univ-lr.fr}\\
}
\maketitle

\abstract{
3D convolutional networks is a good means to perform tasks such as video segmentation into coherent spatio-temporal chunks and classification of them with regard to a target taxonomy. In the chapter we are interested in the classification of continuous video takes with repeatable actions, such as strokes of table tennis. Filmed in a free marker less ecological environment, these videos represent a challenge from both segmentation and classification point of view. The 3D convnets are an efficient tool for solving these problems with window-based approaches. 
}

 

\section{Introduction} 
\label{chap10:sec:intro}

Movement is one of the most important aspects of visual perception, and stimuli associated with movement tend to be biologically significant \cite{MotionBrain}. Psycho-visual experiments show that a complete and exhaustive measurement of a scene is not always necessary to interpret its content. In the animal world, movement information, even partial, may be sufficient to recognize potential food: a frog can distinguish the flight of a fly from a falling leaf without needing to geometrically reconstruct the whole scene.
\par
An illustration of recognition through movement by human beings is provided with the experiments of Johannsson (\textit{Moving Light Display experiment}~\cite{Joh73}) in which the only source of information on a moving actor is given by bright spots attached to a few joints. People shown a static image can only see meaningless dot patterns. However, as soon as they are shown the entire sequence of images, they can recognize characteristic actions such as running or walking, and even the male or female gender of the actors. Such abilities suggest that it is possible to use movement as a means of recognition by itself. \\
In the field of computer vision, describing a video scene through natural language involves the narration of some key events or actions occurring in the scene. Movement analysis is obviously a central element in order to be able to take full advantage of the temporal coherence of the image sequences. Moreover,~\cite{laptevhdr} states that on average, on each video from the web, 35\% of the pixels represent humans. Describing an image sequence from the human movements and activities performed is hence potentially discriminant and relevant for analyzing or indexing videos. These different elements, added to the complexity of the task, highlight why human action recognition in videos has been in past years a very active research topic in computer vision.
Nevertheless, as shows the already large history of research for solving the problem of action recognition in video, when a real-world video scenes have to be analysed~\cite{Stoian16}, motion characteristics alone are not  sufficient. This is why the approaches using both temporal information expressed in terms of motion characteristics, such as velocity field, i.e. Optical Flow (OF) and spatial characteristics derived from colour pixel values in video frames have shown better performances. This is hold both for methods on the basis of hand-crafted features or the ever-winning Convolutional Neural Networks (CNNs).  \\

Focusing on Deep Learning approaches, it is nevertheless interesting to show the problem in its historical perspective for better predicting the future. In this chapter, we will briefly present approaches for action recognition with different features going from handcrafted to produced by Deep NNs. We will speak about the evolution of datasets for development and testing action recognition methods, and introduce recently  created dataset in our research for fine-grained classification of Table Tennis strokes. We will also present recent contributions in fine -grained action recognition with Twin Spatio-temporal networks.

\section{Highlights on Action Recognition Problem} 
\label{chap10:sec:relatedwork}

The problem of recognition of human actions in video has a wide-range of applications, this is why the history of research is quite long-term one. From historical perspective one can distinguish two main approaches with progressive combination of them. The first one consists in designing the so-called "handcrafted" features, expressing characteristics of video frames in a local or holistic manner thus forming a new description space. Classification of features in this new space with machine learning approaches brings the solution to the action recognition problem. The second approach consists in the "end-to-end" solution with Deep Neural Networks which extract features and then classify them.

\subsection{Action classification from videos with handcrafted features}
\label{chap10:subsec:general}

Handcrafted features extracted from videos development started to our best knowledge from feature extraction from images. Efforts were afterwards made for extracting information from the temporal domain. Such features were mainly used in the action recognition task, but also for other tasks such as scene and event recognition from videos. Most of the approaches using handcrafted features seek for their compact representation. The model of Bag of Words (BoW) or Bag-of-Visual Words (BoVW)~\cite{Csurka04} was introduced which allowed for quantizing a large amount of  feature vectors into a set of classes - words of a dictionary of a predefined size.  The classes-words of the dictionary were built by statistical clustering methods such as K-means~\cite{Macqueen67} or more sophisticated vector quantization techniques. This final descriptor of the images, areas in them or video frames were thus a statistical model - a histogram of class - occurrence of the ``words'' in the image. This model, for a long period was used for action recognition, and scene classification in video.

The use of temporal dimension of video with regard to static images was first introduced in 2003 with Spatio-Temporal Interest Points (STIP) features~\cite{Features:LocalSpaceTime}. They were an  extension of the 2D  corner detector to the temporal dimension for video. The equivalent to image corners in video are points which change direction over space and time.
The authors of \cite{Features:LocalSpaceTime} show that their descriptor matches with the action performed, meaning that STIPs will be located where and when the action happens. They apply their method on their newly created \texttt{KTH} dataset, which became one of the first widely used action datasets.  Comparison is done using Support Vector Machine (SVM) classifier and Nearest Neighbor Classification (NNC) on BoW of their spatio-temporal local features. It results in a good classification score for actions which are not similar; but scores on similar actions such as ``Walking'', ``Jogging'' and ``Running'' remained low. The motion in those actions is very similar and the STIPs are concentrated on the same body parts.

\par

The authors of \cite{Dataset:CoffeAndDrinking} use jointly the Histogram of Oriented Gradients (HOG) descriptor and Motion Boundary Histogram (MBH) descriptors, using AdaBoost algorithm for recognition of two actions - smoking and drinking in their own dataset based on movies. At the same time,~\cite{Features:SpaceTimeShape} introduce Space-Time Shapes (STS) features for action classification, along with the new Weizmann action dataset. Classification of STS is based on the same ideas as image shapes classification using Poisson equation. The authors classify the computed shape with Nearest Neighbor Clustering with euclidean distance. They reach an accuracy of $97.8$\% on their dataset with however a low confidence on similar actions. It is important to stress that their dataset is acquired in a controlled environment and has the same complexity as the KTH dataset. 

\par
In \cite{Features:ActionSnippets}, the authors concatenate Gabor filters features and OF features in order to perform action classification with SVM classifier on KTH and Weizmann datasets. They show the superiority of their model on both datasets, compared to other methods, and come to the conclusion that one frame is enough to get a satisfactory classification score. Indeed, using only one frame and the posture of the person, actions in both datasets are easily distinguishable.

\par
In $2009$, action recognition methods were already reaching very high accuracies on both KTH and Weizmann datasets and the introduction of UCF11 dataset gave more space for improvements. Indeed, this dataset is more challenging since it is recorded ``in the wild'', that is in natural conditions, under the constraints of camera motion and flickering for example. The UCF101 \cite{Dataset:UCF101} samples are extracted from the YouTube platform. Along with their dataset, the authors propose a method for classification based on motion features and static features. They use AdaBoost learning method on the histogram-based representation and compare it with k-means clustering method. AdaBoost leads to better results and the hybrid combination resulted in $93.8$\% of accuracy on KTH dataset against $71.2$\% on UCF11 showing the higher complexity of the task for such dataset with the same number of classes.

The authors of \cite{Features:HOF} introduce at the same moment Histogram of Oriented optical Flow (HOF) features and reach $94.4$\% of accuracy on Weizmann actions dataset. The method is simple and easy to reproduce and will be used later on in \cite{Features:DenseTraj}, along with MBH, HOG features to compute dense trajectory features on the basis of dense optical flow field.

\par
In \cite{Features:IDT}, the authors improve dense trajectory features by considering camera motion. Camera motion is estimated using dense optical flow and Speeded Up Robust Features (SURF) descriptors. An homography is estimated using Random Sample Consensus (RANSAC) algorithm. The Improved Dense Trajectories (IDT) yield $91.2$\% of accuracy against $88.6$\% with regard to the original dense trajectory features. This work is used later in  many applications such as action localization. The latter problem consists in defining not only temporal boundaries of actions in the video, but also spatial locus of them. 

\par
In \cite{Localization:Actoms}, the authors redefine actions as ``actoms''. Actom is a short atomic action with discriminative visual information, such as opening a door. It is therefore useful for action localization but can also be applied to classification-by-localization. The definition of actoms is important in the field of action recognition to decompose an action in individual parts. Actoms thus can be present across different actions such as entering or leaving a room with ``opening door'' as an actom, for example. Their understanding can lead to better video representation and accordingly to a better classification. However, a too great number of actoms might lead to teh situation when they are  well presented in the training set, but they would not be present in the test set and be unrelated to the action performed. The number of actoms to consider becomes then a variable to control according to the complexity of the actions to classify. 

\par
Another way to perform  action recognition is developed by~\cite{Localization:TubeletsMotion} who introduce the concept of  Tubelets. It is a sampling method to produce 2D+T sequences of bounding boxes where the action is localized. This method, which tackles the localization and classification problem of actions at the same time, is based on super voxel generation through an iterative process using color, texture and motion to finally create tubelets. Those tubelets are then described by MBH features, and one BoW per class method is used for classification. The classifier with the maximum score assigns the class to the tubelet.


\subsection{The move to DNNs in action recognition problem}
Deep Neural Networks have very quickly outperformed all handcrafted feature-based methods, due to the strong generalization capacity of these classifiers. The method described in~\cite{NN:3DCNN2013}, was one of the first to use Deep Learning via a 3D CNN for action recognition, but they did not obtain better results than the state-of-the-art methods on the \texttt{KTH} dataset. It is only from $2014$ and the innovative \textit{two-stream network} approach of~\cite{NN:SimonyanTwoStream} that temporal coherence will be exploited in CNN and that Deep Learning approaches will begin to supplant other methods. In the development of Deep Learning Methods for Action Recognition we could observe two trends: Deep Convolutional Neural Networks  (CNNs) and recurrent neural networks (RNN) such as LSTM
briefly presented in Chapter~2. Nevertheless, according to \cite{NN:Laptev18} and the own experience of the authors, these networks are more difficult to train than 3D CNNs integrating spatial information along the time dimension in video. They have also difficulties to handle long term temporal interactions. Hence in this chapter we will not focus on them. 
\par
Better performances of Deep Neural Networks (DNNs) with inherent feature extraction from raw video in the end-to-end training and generalization process does not mean that engineered features have to disappear. On the contrary, human understanding of visual scenes influences the choice of the designed features such as e.g. OF. Still recent work may use handcrafted features as a baseline or fuse them with the deep features extracted by a DNN.~\cite{Features:EngineeredFeaturesUsefull} confirm that SVM does not perform better than a partially retrained Deep Convolutional Neural Network (DCNN) and that the learned features lead to better results than engineered ones; however the fusion outperforms the single modalities. These works confirm the findings of the community: the fusion of multiple features allows improving recognition scores for complex visual or multi-modal content understanding~\cite{Features:FusionJenny}.
Action recognition with Deep CNNs was first fulfilled with 2D CNN architectures. Here we will briefly discuss some of them. 
\subsection{2D Convolutional neural Networks for action classification}
2D convolution refers to the fact that convolutions are performed on a 2D spatial support of the image. For RGB data, 2D convolution actually uses 3D kernels to weight each color channel differently. However the way that the kernel will move in the image will only be in 2 dimensions.

\par
In the scope of action recognition, it is  what~\cite{NN:SimonyanTwoStream} perform. They introduce a Two-Stream Convolutional Network which takes one single RGB frame for one stream, and for the other stream, several frames of the computed OF. Each stream is respectively called ``Spatial stream ConvNet'' and ``Temporal stream ConvNet''.

They notice that ``Temporal stream ConvNet'' reaches much better performances compared to ``Spatial stream ConvNet''. This could easily be explained, as in addition to the dynamic genre of the input data, the temporal stream uses up to ten frames against only one for the spatial stream branch. Performances are much more alike when the temporal branch uses only one frame. Of course, the fusion of the two streams using a SVM method performs the best.



\par
An action can be considered as a volume, i.e. tube in the video. This concept of tube - which expresses a homogenous content, a singular action in our case, is used in~\cite{NN:TCNN:Li}.  A Tube Convolutional Neural Network (T-CNN) is introduced which can be sequentially decomposed into two distinct networks. They first create motion-segmented tubes using a Residual Convolutional Neural Network (R-CNN)~\cite{Segmentation:FasterRCNN}. Then those tubes feed a VGG-like network using 20 motion amplitude frames distributed along the channel dimension. 
\par
Until now we were speaking about CNNs. Nevertheless, temporal (recurrent) neural networks also represent an alternative to CNNs with windowing approaches. The focus of our work is on CNNs hence we will just very briefly mention them.  

\par
Long-term Recurrent Convolutional Network (LRCN) models are introduced in~\cite{NN:LRCN} for action recognition. The authors extract features using 2D CNN for each image which feed a LSTM from start to end. The decision is based on the average score. This simple model is tested with OF and a single RGB image. The fusion of the two modalities perform obviously the best.
\par
Numerous works use models based on temporal networks such as Recurrent Neural Network (RNN) and Long Short-Term Memory (LSTM)~\cite{DBLP:journals/access/UllahAMSB18}. However, RNN may be harder to train depending on the application context. Besides, LSTMs are more efficient when they are coupled to the output of a CNN~\cite{NN:LSTM_CNN}.
\par

Now coming back to the convolutional NNs for action recognition, we can satate that 2D CNNs are often used as feature extractors 
and 3D convolutions are performed on the extracted features. Thus the temporal dimension is taken into account. This leads us to now focus on 3D CNN based methods for action recognition.

\subsection{From 2D to 3D ConvNets in Action Classification} 

We can consider videos as 3D data with the third dimension along the time axis and either process them similarly to 2D images, or treat the temporal dimension differently, or extract temporal information such as dynamic data that can feed a DNN.

\par
However, most methods need to consider extra information, which obviously leads to larger networks, greater number of parameters and the need of a greater number of GPUs with stronger capacities. This might not be possible for every research team, and brought some of them to try attaining accurate results with restrictions on the model size or computation time. This aspect thus brings many shades in the performances, and has brought to light many different methods which shall not be compared only in terms of performances, but also by their means to achieve them. In addition to such limitations, the choice of the architecture for a specific task remained open, leading to numerous implementation attempts. 

 3D convolutional neural networks are a good alternative as well for capturing long-term dependencies~\cite{NN:I3DCarreira}, and involve 3D convolutions in space and time. 
 When doing 3D convolution with $C_{in}$ channels on a signal of depth~$D$, width~$W$ and height~$H$, the output value of the layer with input size ($C_{in}$,~D,~H,~W) and output ($C_{out}$, $D_{out}$, $H_{out}$, $W_{out}$) can be precisely described as: 
 \begin{equation}
    out(C_{out_j}) = bias(C_{out_j}) + \sum_{k=0}^{C_{in}-1} weight(C_{out_j},k)\star input(k)
\end{equation} 
where $C_{out_j}$ is the $j^th$ output channel, and $\star$ is the valid 3D cross-correlation operator.
 
\subsection{3D Convolutional neural Networks for action classification}


The Convolutional 3D (C3D) model~\cite{Learning3DCNN} consists of eight consecutive convolutional layers using $3\times3\times3$ kernels and $5$ max-pooling layers. In \cite{NN:TCNN:Hou}, they use this model in a two stream T-CNN. Here videos are first divided into clips of equal length and are segmented using 3D R-CNN to create tube proposals. Tubes are then classified and linked together. By using the features extracted from the segmented video tubes with C3D model, they increase the performance compared to a direct application of the C3D model.

\par
The authors of \cite{NN:TwoStreamResNet} extend what was done in 2D by~\cite{NN:SimonyanTwoStream}, into 3D to introduce their Spatio-Temporal ResNet (ST-ResNet). They replace simple CNN branches by R-CNN with one connection between the two branches. Their results prove that RGB stream processed alone gets better performances than the OF stream. By processing them together, they reach an accuracy of $93.4$\% on UCF101 dataset.

\par
A major breakthrough was proposed by the method of~\cite{NN:I3DCarreira}, with much higher scores obtained on action classification. They present their Two-Stream I3D model as the combination of RGB-I3D and Flow-I3D models trained separately. Each of their models uses inflated inception modules, inspired from the 2D inception modules~\cite{NN:GoogLeNetInception}. The major strength of their model is the pretraining on ImageNet first and then on Kinetics-400 dataset~\cite{Dataset:Kinetics}, more complex than UCF101 with $400$ classes. By using Kinetics, they boost performances from $93.4$\% to $98$\% on UCF101. They reach also $74.2$\% of accuracy on Kinetics-400 dataset. Inception modules have already proven their efficiency for image classification on ImageNet dataset and thus have been sucessfully used in the I3D models.

\par
Long-term Temporal Convolutions (LTC) CNN were introduced by~\cite{NN:Laptev18}. They experiment different temporal sizes for input video clips, in order to improve  classification. Better accuracies are obtained when considering a greater number of frames as input, especially on long-lasting actions which have a longer temporal support.

\par
\cite{NN:TwoStream3D} introduce Spatial-Temporal Pyramid Pooling Layer (STPP) using 3D convolutions in a two-stream like network fed by RGB and OF streams. The output becomes the input of a LSTM network. The use of LSTM allows classification of videos of arbitrary size and length. Each modality performs similarly: $85$\% and $83.8$\% of accuracy for RGB and OF stream respectively. When fused together, the method reaches $92.6$\% of accuracy.

\par
\cite{Attention:BERT} introduce the Bidirectional Encoder Representations from Transformers (BERT) layer to better make use of the temporal information of BERT’s attention mechanism firstly used for language understanding~\cite{Attention::translation}. The BERT layer is based on the use of the Multi-Head Attention layer, which comprises a  Scaled Dot Product layer. The Multi-Head attention layer is a part of a bigger network, the Transformer model which is dedicated to translation tasks. The incorporation of the BERT layer in the REsNeXT, R(2+1)D and I3D models, previously described, improve their performances.
They reach the state-of-the-art results on UCF101 dataset with $98.7$\% of accuracy using the R(2+1)D architecture~\cite{NN:R21D}. It is a ResNet-type architecture with separable temporal and spatial convolutions and a final BERT layer in order to better use the obtained features. One important point to stress is also the use of IG65M dataset~\cite{DatasetPretrain:IG65M} for pre-training the model. IG65M dataset is build from the Kinetics-400~\cite{Dataset:Kinetics} class names. Those class names are then used as hashtags on Instagram and lead to $65$M clips from $400$ classes. Their dataset is however not publicly available.

\subsection{Video understanding for racket sports}
\label{chap10:subsec:sport}

Our interest is fine-grained action recognition with application in table tennis. We therefore present methods focusing on video classification and/or segmentation in the domain of racket sports.

\cite{Dataset:MotionForBalletFootTennis} propose a motion descriptor based on optical flow in order to classify actions in sports. For this purpose they consider $3$ different datasets: Ballet, Football and Tennis datasets. They track the player (or the person performing the action) and build a 3D volume based on their motion. Their motion descriptor has $4$ channels: the positive and negative values for horizontal and vertical motions.
\par
Then classification is performed following a nearest neighbour approach using similarity metric:

\begin{equation}
    S=\sum_{k}C_1(k)C_2(k)
\end{equation}

with $C_1$ and $C_2$ being two cuboid samples based on the motion descriptors at coordinate $k$.

Even if Ballet dataset and Tennis dataset are acquired in a controlled environment, performances for the Tennis dataset are more limited. Football dataset comes from broadcast source which explains the limited performances. Moreover, the number of classes for the Tennis dataset is lesser than the two others, however, it is where their method is the less efficient. This underlines the greater complexity of racket sport and their fine-grained aspect.

\par
Another research field in video classification aims at identifying the different parts of tennis broadcasting. To do so HMMs are applied to tennis action recognition by~\cite{Ewa}. Their model is statistic and integrate the structure of tennis match. They combine audio and key frame features to be able to segment, with a good accuracy, the different parts of the tennis broadcasting such as the first serves, rallies, replays and breaks. The sound of the crowd such as applause, the sound of the ball or the commentator speech combined with key frames which capture visual information lead to $86$\% of segmentation accuracy compared to $65$\% and $77$\% with only respectively visual features and audio features. Such applications are interesting for sport coaches who wish to comment and examine only sequences of sports.

\par
\begin{figure}[htbp]
	\centering
	Global accuracy of $77.6$\%\\
	\includegraphics[width=.48\linewidth]{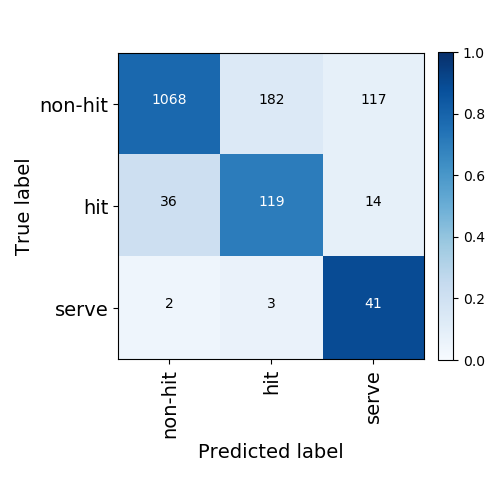}~~
	\includegraphics[width=.48\linewidth]{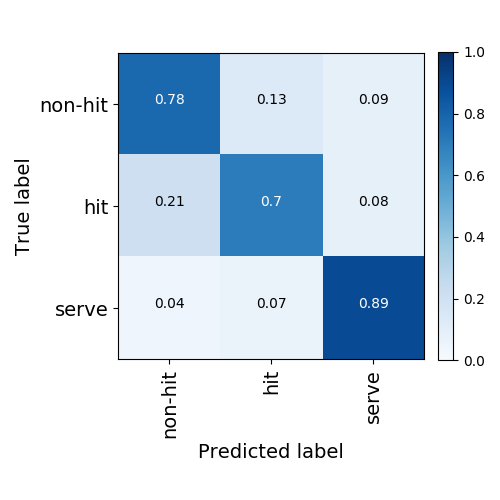}
	\caption{Confusion matrix of tennis game from TV brodcast~\cite{Dataset:ACASVA}. Left with number of samples and right normalized.}
	\label{chap10:fig:confTennis}
\end{figure}

\cite{Dataset:ACASVA} present a new dataset for tennis actions. This one contains only three types of classes: ``hit'', ``serve'' and ``non-hit-class''. The dataset is build from TV broadcasts of tennis games (matches of females in the Australian Open championships). They are interested in action localization and their classification. To do so, they introduce a local BoW method on the Spatio-Temporal gradients HOG3D features which are an extension of the classical 2D HOG features in $3$ dimensions. They also use STS features. Both features are from the located actor and classification is performed using Fisher discriminant analysis. They obtain an accuracy of $77.6$\% using STS model based. Their confusion matrix is represented in figure~\ref{chap10:fig:confTennis}.

One can see that ``serve'' samples are easier to classify than ``hit'' or ``non-hit'' samples. This is certainly due to the time that a service takes and its decomposition in time, which starts by large movement of the player when launching the ball. Hit and non-hit classes are then harder to distinguish because the hit class is very limited in time. Looking only at the player shape, the ball might not be visible, and features might look the same as when the player is simply moving in the field.

\par
Recently, deeply related to our domain,~\cite{TT:OFSingularitiesForStrokeClassification} use the OF Singularities with BoW and SVM in order to classify very similar actions. This task is also called fine-grained action classification. They apply their method on the \texttt{TTStroke-21} dataset which contains $20$ different strokes and a negative class. Their method is inspired from~\cite{Features:FrequencyOFCrititicalPoint} which uses the trajectory of critical points for classification. In this case, the actions to recognize are the different types of strokes preformed during table tennis training session. However, the scores remain low due to the high similarity of the different strokes and the limited amount of video samples. It makes generalization of extracted features harder. 

\par
Table Tennis stroke recognition is also performed by~\cite{TT:RecognitionWithSensor}. It is based on their Body Sensor Network (BSN). Their sensors collect acceleration and angular velocity information from the upper arm, lower arm and the back of the player. From the recorded signals, they extract Principal Component Analysis (PCA) features which are then fed to a SVM. They reach an accuracy of $97.4$\%, however they use only $5$ classes: ``forehand drive, ``block shot'', ``forehand chop'', ``backhand chop'' and ``smash''. Similarly,~\cite{TT:RecognitionWithSensorHybrid} recently proposed classification from integrated wearable sensors using K-means and DBScan clustering. Their taxonomy is more limited than in \texttt{TTStroke-21} since they use only $9$ classes across badminton and table tennis sports. They reach an accuracy of $86.3$\% when considering all the classes. This score reaches $92.5$\% when considering only table tennis but this classification is limited to $4$ classes: ``Service'', ``Stroke'', ``Spin'' and ``Picking up''. The extent of their taxonomy is thus limited and does not contribute much to the player experience. Furthermore, using such sensors, strongly limits the application possibilities and has a greater cost regarding training equipment adaptation. Also their system does not offer visualisation of the stroke performed since it is based on sensors, and it limits the feedback for the player.

\par
Recently, a method is introduced by~\cite{TT:Tac-Simur} to get the tactics of the players based on their performance in past matches. Their model is based on Hidden Markov Model (HMM) and aims at characterizing and simulating the competition process in table tennis. They use richer taxonomy and terms of stroke techniques than the previously presented methods : $13$ different classes and $4$ player positions which can be combined. Compared to \texttt{TTStroke-21}, we consider $10$ classes with $2$ player positions: ``Forehand'' and ``Backhand''. Their goal is therefore, not to classify an input, but to simulate matches between two different players. It is not directly linked to action recognition methods, but it does give a tool for players to simulate sport encounters and give credits to the \texttt{TTStroke-21} dataset which propose much richer taxonomy than previous datasets.

\par
Thus having analysed the two kinds of approaches: with handcrafted features and with Deep Neural Networks we can state the following. Despite the use of temporal information coming from OF and derived features, the methods with handcrafted features allow a good classification on datasets that remain simple: either with a low number of classes or with classes that are easily separable. It does become more complicated when the task focuses on one particular sport with different actions within or in case, when the complexity of ``in-the-wild'' scenes is higher. Here, already earlier research work have given a direction to follow: the use of Deep Neural Networks. 
Before we afford our solutions, it is interesting to review existing datasets which are used by the community for action recognition in video.


\section{Datasets for action recognition}

The need of datasets for action recognition has grown those last years. These datasets can change in terms of number of videos starting from a few videos up to millions of them. In addition to their size, the number of categories and their complexity also vary from few classes up to few hundreds, or even thousands of them. Each dataset can be labelled with annotations either by enriching the terminology, localising the action in space and time or by adding modalities information such as skeleton. 


\subsection{Annotation processes}
\label{chap10:subsec:Annotation_process}

One can distinguish two ways to annotate a dataset: automatic or "by hand", that is by a human operator. Label propagation techniques were an attempt to leverage the gap between fully ``manual'' annotation and automatic one~\cite{LabelProp13}. Here the dataset is represented as a big graph with samples-nodes and associated similarity metric between them. The label propagation is fulfilled from manually annotated nodes to unlabelled ones by optimal search on graphs. Nevertheless, the practices in annotation of video datasets are such that, the automatic annotation by a concept recognition method at hand is a first step and then human intervention is required to filter out automatic annotation errors. Also, it is common to have a dataset split into ``auto''  and ``clean'' sets. The ``auto'' being the one annotated using automatic methods and ``clean'' the one automatically labeled, verified and adjusted by hand. The two annotation processes are first described before presentation of the datasets.

\subsubsection{Automatic annotation}

Tags from social network platforms can be used to collect rich datasets. It is the case, for example, with the IG65M dataset~\cite{DatasetPretrain:IG65M}. To collect it, the authors use Kinetics-400~\cite{Dataset:Kinetics} dataset class names as hashtags on Instagram and collect $65$M clips from $400$ classes. Such annotation process requires filtering in order to refine the annotations.

\par
In movies, the script can also help to automatically label the sequence. It is for example what~\cite{Dataset:Hollywood2} have done for the Hollywood2 dataset. They generate the samples this way and clean them manually for the test set. Similarly, datasets can be constructed from the description of the videos from online platforms hosting them. Then, according to a description, datasets can be generated in an automatic way. 

\subsubsection{Manual annotation}

The most common way to annotate a dataset, especially when it is not large, is to label all the samples by hand. Some tools might be used to help in the process such as a pre-classification if a model already exists, or localization segment candidates of the actions when the video is untrimmed. One can distinguish two ways in the hand-labelling process: if the annotation is done by one person or several. By one person, the risk is that an inattention might lead to errors in the dataset or make it biased according to the point of view of the annotator. To overcome this issue, a crowdsourcing method can be used. 

\par
Crowdsourcing is based on the annotation of the same segments by different persons. It relies on the collective intelligence and should give better results than with only one person annotating. Outliers in annotations are not considered in the final decision. Different rules might apply, e.g take the mean of the annotators when possible or consider only the annotator that performed the best today; in order to take an annotation decision. There are also datasets which provide gross croudsourced annotations and it is the team working on the dataset that decides which decision to make.

\par
A new trend appeared recently: the use of Amazon Mechanical Turk (AMT) \footnote{\url{https://www.mturk.com/}}. AMT, also called ``MTurk'', is a crowdsourcing marketplace for individuals and businesses to outsource their processes and jobs to a distributed workforce who can perform these \textit{micro-tasks} virtually. Here it is applied to annotation and AMT are paid according to the number of annotations performed. It started to be used with ImageNet dataset dedicated to image classification with $3.2$ millions of images over $5~247$ classes. The strategy is two folds:

AMT workers verify the presence of the action in the video candidates and they can also temporally annotate them. It is often coupled with a crowdsourced method meaning that each video will be annotated by several AMT workers. This allows the construction of a large dataset in a short amount of time. 

\subsection{Datasets for action classification}
\label{chap10:subsec:action_datasets}

Datasets of actions classification problem can be categorized in many ways. In this section, the datasets are grouped according to the acquisition process: acquired in a controlled environment, extracted from movies or recorded ``in the wild''. Obviously, such a categorization is not perfect since some datasets mix different types of videos. Without being exhaustive we will present some of them focusing mainly on sport video datasets.

\subsubsection{The acquisition-controlled datasets}

These datasets are often self made by the authors who decide in what type of environment the actions will be performed. It does not always mean that the dataset is easier than ``in the wild'' since difficulties can be added on purpose. The databases from broadcasts or recordings not meant for action recognition task are also considered in this subsection because the acquisition environment can be taken into account in the classification process.\\

\textbf{KTH and Weizmann datasets.} These two datasets were the most popular at the early ages of action recognition research. Despite their simplicity, some researchers are continuing  using them as a benchmark.

KTH introduced in~\cite{Features:LocalSpaceTime}\footnote{\url{https://www.csc.kth.se/cvap/actions/}} stands for ``Kungliga Tekniska Högskolan'' in Swedish which is the Royal Institute of Technology (Stockholm, Sweden), institution of the authors. The dataset is composed of $6$ classes: ``Walking'', ``Jogging'', ``Running'', ``Boxing'', ``Handwaving'' and ``Hand clapping''. The acquisition was done in a controlled environment, with a homogeneous background, static camera at $25$ fps, with $25$ actors and has $2~391$ video clips across $600$ videos. Videos are recorded outdoors and indoors. Weizmann dataset\footnote{\url{http://www.wisdom.weizmann.ac.il/~vision/SpaceTimeActions.html}}~\cite{Features:SpaceTimeShape} is quite similar but is enriched with more actions, e.g. `jumping- jack'', ``galloping-sideways''... increasing the number of action classes to 9. It is constituted of $81$ video sequences recorded at $25$ fps at low resolution $(180\times144)$. 
\par
The necessity of action recognition "in-the-wild" yielded production of much more complex datasets described below.

\vspace{0.2cm}
\textbf{ACASVA.} \cite{Dataset:ACASVA} ``Adaptive Cognition for Automated Sports Video Annotation'' (ACSAVA) introduce a tennis action dataset \footnote{\url{https://www.cvssp.org/acasva/Downloads.html}}. Their objectives is to evaluate classical action recognition approaches with regard to  player action recognition in tennis games. The data are collected from tennis TV  broadcasts. The videos are then spatially segmented on the players and temporally annotated using $3$ classes: ``hit'', ``serve'' and ``non-hit''. The complexity of the the dataset remains simple.


\vspace{0.2cm}
\textbf{FineGym dataset.}
The FineGym dataset\footnote{\url{https://sdolivia.github.io/FineGym/}}~\cite{Dataset:Gym}, is a fine-grained action dataset with a special focus on gym sport. The authors use a rich taxonomy to decompose each actom of structured gymnastic figures. They use three-level semantics and analyse four different gymnastic routines: balance-beam, uneven-bars, vault and floor exercise. They have a total of $530$ element categories but only $354$ have at least one instance. This rich amount of categories is due to all the combinations of possible actoms. The authors offer two settings: Gym288 with $288$ classes but of very unbalanced distribution and Gym99, more balanced but with ``only'' $99$ classes. The total number of samples considering all classes reaches $32~697$. The $708$ hours of videos are hosted on YouTube with most of them in high resolution.

\vspace{0.2cm}
\textbf{TUHAD.} The TUHAD dataset~\cite{Dataset:Taekwondo} is also a dataset dedicated to fine-grained action recognition on Taekwondo sport. The dataset was recorded with the help of $10$ Taekwondo experts using two Kinect cameras with front and side view. The number of classes is low: with only $8$ Taekwondo moves. $1~936$ action samples are recorded with depth and IR images along with the RGB data.  The classes are very similar in many ways but a foot position, which might be overcome with proper features.

\subsubsection{Datasets from movies}

Despite these datasets do not generally contain sport actions, they are also interesting as they comprise recordings of natural behaviour of actors in cluttered environments.
Thus Drinking and Smoking Dataset~\cite{Dataset:CoffeAndDrinking} is composed of sequences from Jim Jarmush Movie ``Coffee and Cigarettes''. It was designed for joint action detection and classification for the $2$ classes: ``smooking'' and ``drinking'' with respectively $141$ and $105$ samples.
\vspace{0.2cm}
\par
\textbf{The Hollywood2 dataset}\footnote{\url{https://www.di.ens.fr/~laptev/actions/hollywood2/}}~\cite{Dataset:Hollywood2} was designed for action and scene classification. It contains $10$ scene classes and $12$ actions: ``Answer phone'', ``Drive car'', ``Eat'', ``Fight person'', ``Get out car'', ``Hand shake'', ``Hug person'', ``Kiss'', ``Run'', ``Sit down'', ``Sit up'' and ``Stand up''; over $7$ hours of video from $69$ movies. They have a total of $1~694$ actions samples. The difficulty lies in the fact that different actions can happen in the same sequence.

\subsubsection{In-the-wild datasets}

``In-the-wild'' means that the videos are from different sources and can be recorded by professionals or amateurs. They thus may contain camera motion, strong blur, occlusions... everything that can make the action recognition task harder. However, videos can also contain much background information, which might be an exploitable source for training of classification models.

\vspace{0.2cm}
\textbf{The UCF datasets.}
The UCF datasets~\cite{Dataset:UCF101} have become very popular for developing and benchmarking methods for action recognition in sport video. UCF title comes from the name of the university in which the datasets have been developed: University of Central Florida.
\par
The first UCF dataset was UCF-Sports. It contains various sequences from broadcast TV channels across $9$ different sports:  ``diving'', ``golf swinging'', ``kicking'', ``lifting'', ``horseback riding'', ``running'', ``skating'', ``swinging a baseball bat'', and ``pole vaulting''. Pole vaulting is split in $2$ classes: ``Swing-Bench'' and ``Swing-Side'' totaling $10$ classes. It first contained $200$ sequences (reduced to $150$ later) with an image resolution of $740\times480$ at $10$~fps.
\par
Later, the UCF YouTube Action also called UCF11 dataset\footnote{\url{www.crcv.ucf.edu/data/UCF\_YouTube\_Action.php}} is introduced. It consists of $11$ classes from $1~160$ videos from the online video platform YouTube.
\par
UCF50 is an extension of UCF11 with a total of $50$ action classes. This version is then extended to make UCF101 dataset\footnote{\url{www.crcv.ucf.edu/data/UCF101.php}}. UCF101 includes a total number of $101$ action classes which can be divided into five domains: ``Human-Object Interaction'', ``Body-Motion Only'', ``Human-Human Interaction'', ``Playing Musical Instruments'' and ``Sports''. Constructed from $2500$ videos ``in-the-wild'', they extract a total of $13~320$ clips in order to have at least $101$ clips per class. The dataset is widely used by the scientific community and led to the THUMOS challenge\footnote{\url{www.thumos.info}} held in $2013$, $2014$ and $2015$. The dataset was cleaned and enriched with temporal annotations in $2015$ in order to provide qualitative benchmark for different methods and be used also for spatio-temporal localization and temporal detection only.

\vspace{0.2cm}
\textbf{The Kinetics datasets.}
The kinetics datasets\footnote{\url{https://deepmind.com/research/open-source/kinetics}}: Kinetics-400~\cite{Dataset:Kinetics} , Kinetics-600~ and Kinetics-700~\cite{Dataset:Kinetics700} consider respectively $400$, $600$ and $700$ action classes. They are all financed by DeepMind company, specialized in AI which, from $2014$, belongs to Google. In the taxonomy of actions they contain, we find sport actions as well, e.g. "playing squash or racquetball". 

\par
The videos are collected from YouTube video platform, automatically annotated and candidates are refined using AMT. The difference between the versions of the datasets lies in:
\begin{itemize}
    \item the number of classes: the number of classes has increased over time. New classes were added and pre-existing classes were refined. Some were merged.
    \item the amount of videos: the number of videos started with $306~245$ clips and more than doubled in the last version
    \item the splits between the different sets: training, validation and test sets have been modified over time. For example, samples belonging to the training set in the first version might belong to the test set in the last version.
\end{itemize}

\vspace{0.2cm}
\textbf{AVA and AVA-kinetics.}
In \cite{Dataset:AVA}, the authors introduced the AVA dataset\footnote{\url{https://research.google.com/ava/}} in order to perform joint localization and classification of actions. It contains $437$ videos gathered from YouTube, $15$ minutes are extracted from them and annotated every second. They use a vocabulary of $80$ atomic actions. The difficulty in this dataset is the overlapping actions in time and their localization. They offer a split of the dataset by extracting $900$ video segments of $3$~seconds from all the $15$~minutes videos. By doing so, the $55$~hours of video are split in $392~000$ overlapping segments. 
The AVA-Kinetics dataset \cite{Dataset:AVA_Kinetics}\footnote{\url{https://deepmind.com/research/open-source/kinetics}}. Is the merger of the two: Kinetics-700 and AVA datasets. Kinetics videos were annotated using AVA protocol. The dataset thus contains over $230~000$ video clips spatially and temporally annotated using the $80$ AVA action classes.

\vspace{0.2cm}
\textbf{SAR4.} \cite{Sport:RecognitionSoccerPose} present the SAR4 dataset which focuses on action in football sport (or soccer). They track and label the players from available videos on YouTube. The actions performed by the tracked players are then annotated using a taxonomy of $4$ classes:  ``dive'', ``shoot'', ``pass received'' and ``pass given''. The total number of sequences is $1~292$ with actions lasting from $5$ up to $59$~frames.
The discussed datasets are summarized in table \ref{chap10:tab:datasets}

\begin{table}[htbp]
    \centering
    \caption{Presentation of the different action datasets in terms of number of classes, acquisition process, the amount of videos and the number of extracted clips.}
    \begin{tabular}{cccccc}
        Datasets         & \# classes        & Acquisition       & \# videos         & \# clips \\ \hline

        KTH~\cite{Features:LocalSpaceTime} 
                        &$6$                & Controlled        &$600$              &$2~391$    \\
        Weizmann~\cite{Features:SpaceTimeShape}
                        &$9$                & Controlled        & -                 &$81$       \\
        Coffee and Cigarettes~\cite{Dataset:CoffeAndDrinking}
                        &$2$                & Film              &$1$                &$246$      \\
        UCFSports
                        &$10$               & Broadcast         & -                 &$150$      \\
        UCF11
                        &$11$               & In the wild       & -                 &$1~160$    \\
        Hollywood2~\cite{Dataset:Hollywood2}
                        &$12$               & Films             &$69$               &$1~694$    \\

        UCF101~\cite{Dataset:UCF101}
                        &$101$              & In the wild       &$2~500$            &$13~320$   \\
        AVA~\cite{Dataset:AVA}
                        &$80$               & In the wild       &$437$              &$392~000$  \\               
        SAR4~\cite{Sport:RecognitionSoccerPose}
                        &$4$                & Broadcast         & -                 &$1~292$    \\
        Kinetics-700~\cite{Dataset:Kinetics700}
                        &$700$              & In the wild       & -                 & $650~317$ \\
        FineGym~\cite{Dataset:Gym}
                        &$354$              & Broadcast         & -                 &$32~697$   \\
        AVA-Kinetics~\cite{Dataset:AVA_Kinetics}
                        &$80$               & In the wild       & -                 &$230~000$  \\                

    \end{tabular}  

    \label{chap10:tab:datasets}
\end{table}

\subsection{The \texttt{TTStroke-21} dataset}
\label{chap10:subsec:TTStroke}

The \texttt{TT-Stroke21} dataset was recorded for fine-grained recognition of sport actions, in the context of the improvement of sport performance for amateurs or professional athletes. Our case study is table tennis, and our goal is the temporal segmentation and classification of strokes performed. The low inter-class variability makes the task more difficult for this content than for more general action databases such as UCF or Kinetics.

\par
Twenty stroke classes and an additional rejection class have been established based on the rules of table tennis. The filmed athletes are students, and their teachers supervise the exercises performed during the recorded sessions. The objective of table tennis stroke recognition is to help the teachers to focus on some of these strokes to help the students in their practice.

\par
Table tennis strokes are most of the time visually similar. Action recognition in this case requires not only a tailored solution, but also a specific expertise to build the ground truth. This is the reason why annotations were carried out by professional athletes. They use a rather rich terminology that allows the fine-grained stroke definition. Moreover, the analysis of the annotations shows that, for the same video and the same stroke, professionals do not always agree. The same holds for defining temporal boundaries of a stroke, which may differ for each annotator. This variability cannot be considered as noise, but shows the ambiguity and complexity of the data and has to be taken into account. We call this new database \texttt{TTStroke-21}, \texttt{TT} standing for Table Tennis and $21$ for the number of classes.

\subsubsection{\texttt{TTStroke-21} acquisition}

\begin{figure}[hptb]
        \centering
        \begin{tabular}{cc}
              \includegraphics[height=.34\linewidth]{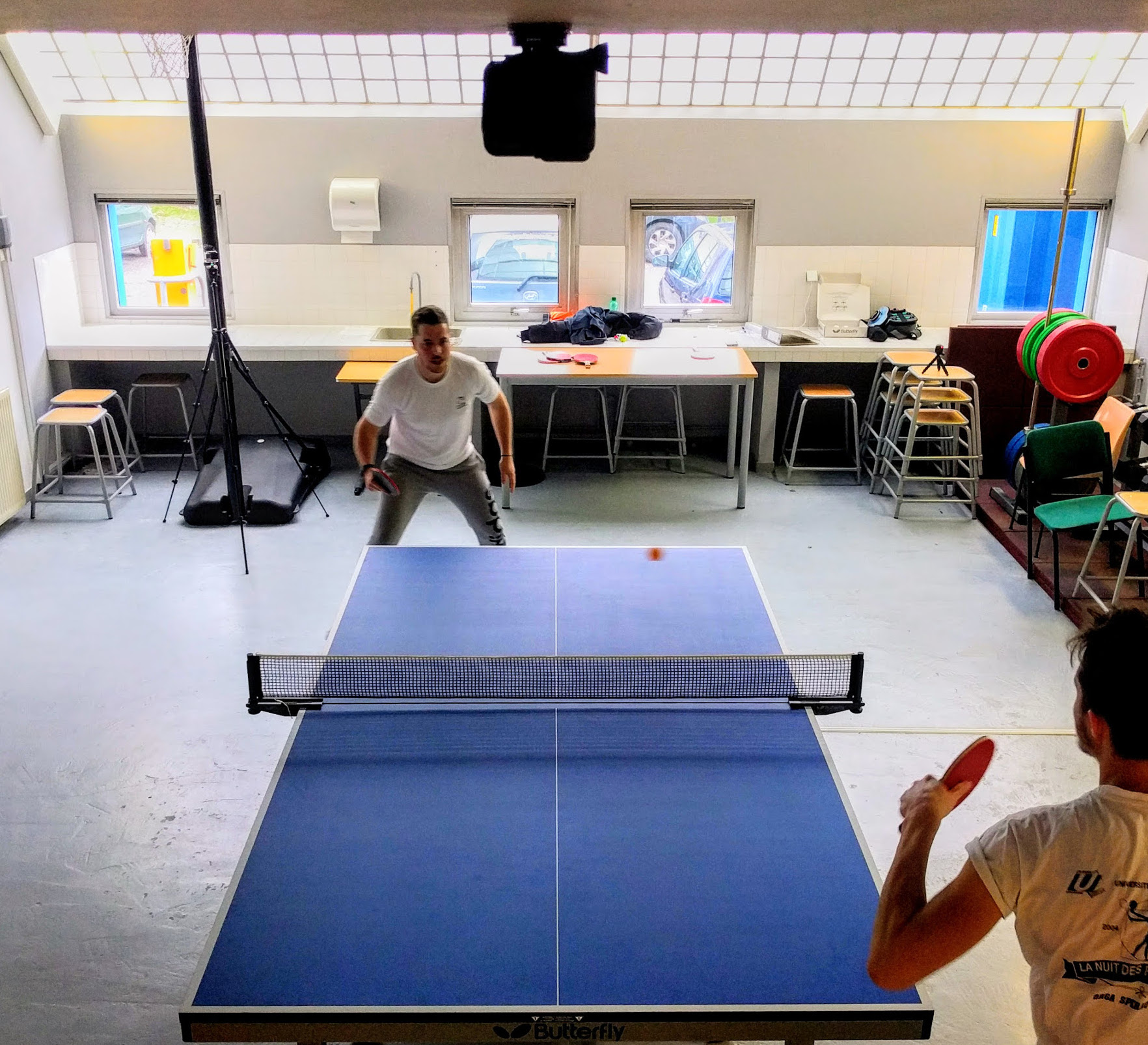}&
              \includegraphics[height=.34\linewidth]{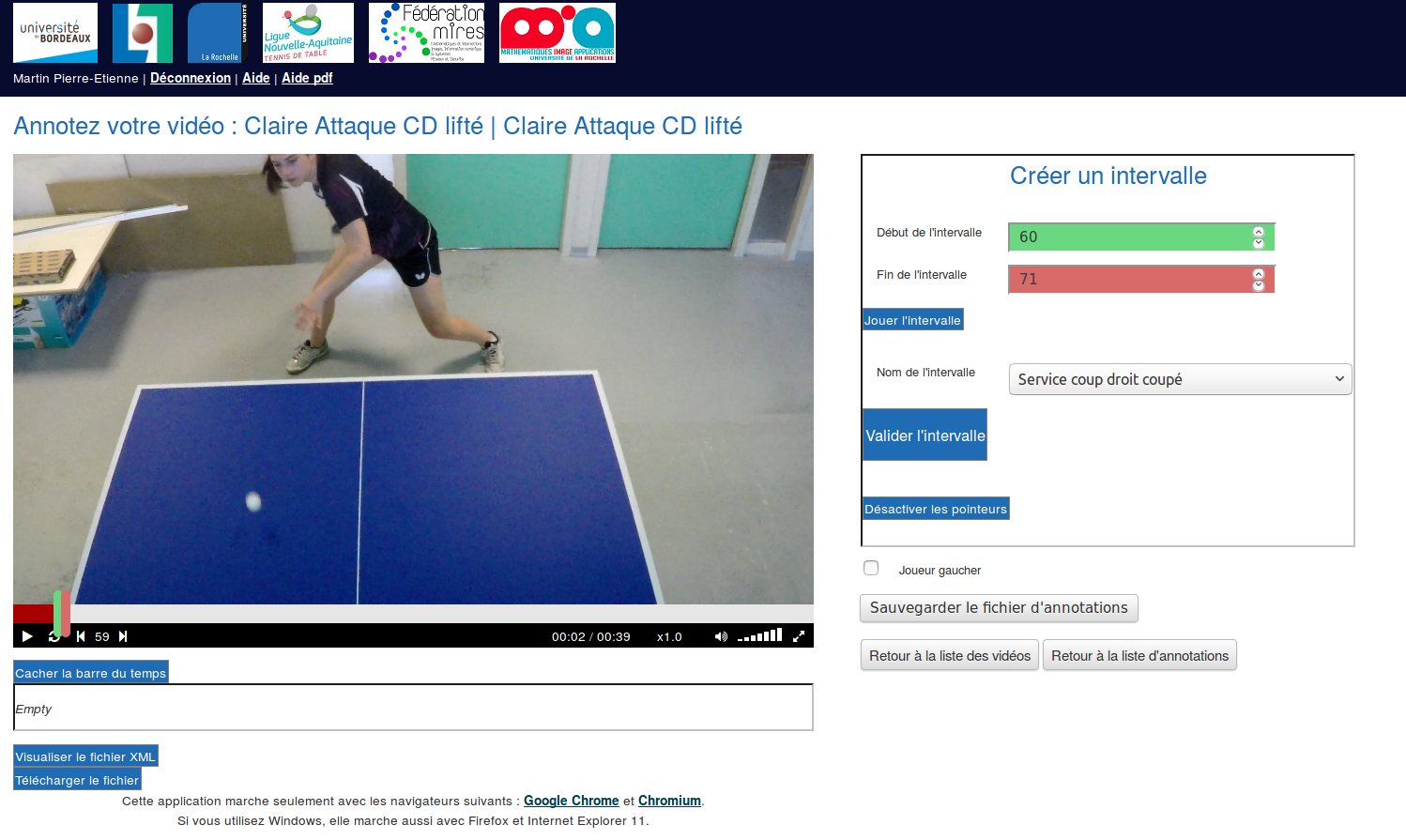}\\
              \textbf{a.} Video acquisition with  &
              \textbf{b.} Annotation platform\\
              aerial view from the ceiling
        \end{tabular}
        \caption{Overview of the \texttt{TTStroke-21} dataset.}
        \label{chap10:fig:acquisition}
\end{figure}

\texttt{TTStroke-21} is composed of videos of table tennis games with $17$ different players. The  sequences are recorded indoors without markers using artificial light and light-weight cameras. The recording setting is illustrated in figure~\ref{chap10:fig:acquisition}.a.

The player is filmed in three situations:

\begin{itemize}
    \item performing repetition of the same stroke. However those repetitions might fail once or several times in the video and the player might do another stroke than the one expected.
    \item simple exchanges between two players: those exchanges are meant to practise the different techniques. 
    \item in match conditions: the players are meant to mark points. The game speed is much faster and strokes are shorter in time.
\end{itemize}

\subsubsection{\texttt{TTStroke-21} annotation}

The annotation process was designed as a crowdsourcing method. The annotation sessions were supervised by professional table tennis players and teachers. A user-friendly web platform has been developed by our team for this purpose (figure~\ref{chap10:fig:acquisition}.b), where the annotator spots and labels strokes in videos: starting frame, end frame and the stroke class. The annotator also indicates if the player is right-handed or left-handed. The taxonomy is built upon a shake-hand grip of the racket leading to forehand and backhand stroke according to the  side of the racket used.

The taxonomy comprises 20 table tennis stroke classes. All the strokes can, as well, be divided in two super-classes: \textbf{Forehand} and \textbf{Backhand}. The linguistic analysis of annotations shows that for the same video and the same stroke, professionals do not employ the same degree of details in their annotations. The same problem occurs with temporal analysis: for instance, a service (first stroke when the player releases the ball) might be considered to start i) when the player is in position, ii) when the ball is released or iii) when the racket is moving. 



\par
Since a video can be annotated by several annotators, temporal annotations needed to be filtered. An overlap between each annotation of $25$\% of the annotated stroke duration is allowed. Above this percentage, the annotations are considered to be part of the same stroke and are temporally fused.

\par
Another filter is applied by checking if labels of the same stroke are consistent. If not, this portion of video is not considered in our classification task. This filtering, based on multiple annotations for the same recorded video, can still leave some labeling errors since multiple labeling of the same clip by different annotators was not always easy to meet in practice.

\subsubsection{Negative samples extraction}
\label{chap10:subsubsec:negative_samples}

Negative samples are created from videos with more than $10$ detected strokes. This was decided after noticing how some videos were poorly annotated and could lead to include actual strokes as negative samples.
\par
The negative samples are video sub-sequences between each detected stroke. We allow the overlap with the previous and the subsequent stroke of $10$\% of our target time window length: $0.83$~seconds, which allows to capture short strokes without considering another one. This represents $100$~frames at $120$~fps.

\subsubsection{Data for evaluation}
\label{chap10:subsubsec:DataForEval}

$129$ videos at $120$~fps have been considered. This content represents $94$ minutes of table tennis games, totalling $675~000$ video frames and $1~387$ annotations. After filtering, $1~074$ annotations were retained. The peak statistics of stroke duration are $min = 0.64$~seconds, $max= 2.27$~seconds and the average duration is $1.46$~seconds with standard deviation of $0.36$. Accordingly, a total of $1~048$ strokes were extracted with a $min$ duration of $0.83$~seconds, a $max$ duration of $2.31$~seconds and an average duration of $1.47$~seconds with standard deviation of $0.36$. Some annotations were merged making the statistical duration a bit longer. After these steps, $681$ negative (non-stroke) samples were extracted. They have a mean duration of $2.34$~seconds and standard deviation of $2.66$~seconds. This high standard deviation comes from the non game activity of long period between strokes, which can be due to a ball lost or talks of players between games. However, as presented in table~\ref{chap10:table:Datasets} representing the distribution over the split of the dataset, not all negative samples are considered to avoid biases in the training and evaluation processes.

\begin{table}
  \centering
  \caption{Datasets Taxonomy of TTStroke-21}
  \label{chap10:table:Datasets}
  \begin{tabular}{cccccccc} 
      \cline{2-8}\noalign{\smallskip}
      & \multicolumn{4}{c}{\textbf{\# Samples}} & \multicolumn{3}{c}{\textbf{\# Frames}}\\
      \noalign{\smallskip}\hline\noalign{\smallskip}
      \textbf{Table tennis strokes} & \textbf{Train} & \textbf{Val} & \textbf{Test} &  \textbf{Sum} & \textbf{Min} & \textbf{Max} & \textbf{Mean*} \\
      \noalign{\smallskip}\hline\noalign{\smallskip}
      Def. Backhand Backspin	&$22$	&$6$	&$3$	&$31$   &$121$  &$233$  &$189\pm25$ \\
      Def. Backhand Block		&$19$	&$5$	&$3$	&$27$   &$100$  &$261$  &$131\pm37$ \\
      Def. Backhand Push		&$6$	&$2$	&$1$	&$9$    &$121$  &$229$  &$155\pm31$ \\
      Def. Forehand Backspin	&$29$	&$8$	&$4$	&$41$   &$129$  &$229$  &$177\pm25$ \\
      Def. Forehand Block		&$8$	&$2$	&$2$	&$12$   &$100$  &$137$  &$115\pm14$ \\
      Def. Forehand Push		&$23$	&$7$	&$3$	&$33$   &$105$  &$177$  &$143\pm19$ \\
      Off. Backhand Flip		&$25$	&$7$	&$3$	&$35$   &$100$  &$265$  &$195\pm49$ \\
      Off. Backhand Hit		    &$28$	&$8$	&$4$	&$40$   &$100$  &$173$  &$134\pm21$ \\
      Off. Backhand Loop		&$21$	&$6$	&$3$	&$30$   &$100$  &$229$  &$155\pm32$ \\
      Off. Forehand Flip		&$31$	&$9$	&$5$	&$45$   &$113$  &$269$  &$186\pm44$ \\
      Off. Forehand Hit		    &$45$	&$13$	&$6$	&$64$   &$100$  &$233$  &$158\pm34$ \\
      Off. Forehand Loop		&$23$	&$7$	&$3$	&$33$   &$101$  &$277$  &$177\pm43$ \\
      Serve Backhand Backspin	&$56$	&$16$	&$8$	&$80$   &$133$  &$261$  &$188\pm31$ \\
      Serve Backhand Loop		&$43$	&$12$	&$6$	&$61$   &$100$  &$265$  &$186\pm42$ \\
      Serve Backhand Sidespin	&$60$	&$17$	&$9$	&$86$   &$129$  &$269$  &$193\pm33$ \\
      Serve Backhand Topspin	&$57$	&$16$	&$8$	&$81$   &$100$  &$273$  &$175\pm48$ \\
      Serve Forehand Backspin	&$58$	&$17$	&$8$	&$83$   &$125$  &$269$  &$182\pm35$ \\
      Serve Forehand Loop		&$56$	&$16$	&$8$	&$80$   &$100$  &$273$  &$171\pm51$ \\
      Serve Forehand Sidespin	&$57$	&$16$	&$9$	&$82$   &$101$  &$273$  &$192\pm39$ \\
      Serve Forehand Topspin	&$67$	&$19$	&$9$	&$95$   &$100$  &$273$  &$184\pm52$ \\
      Non strokes samples		&$74$	&$21$	&$11$	&$106$  &$100$  &$1255$  &$246\pm154$ \\
      \noalign{\smallskip}\hline\hline\noalign{\smallskip}
      \textbf{Total}		&$808$	&$230$	&$116$	&$1154$ &$100$  &$1255$  &$182\pm65$ \\
      \noalign{\smallskip}\hline
  \end{tabular}
  \flushleft
  * in the form: mean value $\pm$ standard deviation
\end{table}

\section{TSTCNN - A Twin Spatio-Temporal 3D Convolutional Neural Network for action recognition} 

After having reviewed a bunch of methods for action recognition in video and of reference datasets, we introduce here our solution to the problem of fined-grained action recognition in video with a 3D CNN we call TSTCNN - a twin spatio-temporal CNN. 

\subsubsection*{A two stream architecture}  

\begin{figure}[ht]
    \centering
    \includegraphics[width=\linewidth]{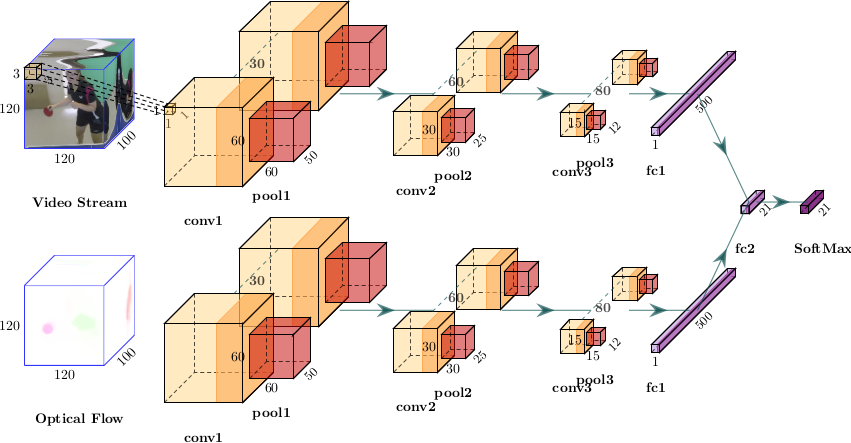}
    \caption{TSTCNN - Twin Spatio-Temporal Convolutional Neural Network architecture. The video stream and its optical flow are processed in parallel through the network composed of successive 3D convolutional layers and 3D pooling layers and the extracted features a merge at the end through a bilinear transformation giving as output a classification score.}
    \label{chap10:fig:architecture}
\end{figure}

The Twin Spatio-Temporal Convolutional Neural Network (TSTCNN), denoted as illustrated in figure \ref{chap10:fig:architecture}, is a two stream 3D Convolutional Network constituted of $2$ branches. Each branch follows the same structure: $3$ blocks constituted of a 3D convolutional layer using kernels of size ($3 \times 3 \times 3$), with stride and padding $1$ in all directions and ``ReLU'' as activation function, feeding a 3D Max-Pooling layer using kernels of size ($2 \times 2 \times 2$) and floor function. From input to output, the convolutional layers use $30$, $60$ and $80$ filters. Each branch ends with a fully connected layer of size $500$. The two branches are combined using a bilinear transformation with Softmax function to output a classification score of size $21$ corresponding to the number of classes considered in our task.

\subsubsection*{Dynamic data}  

The use of dynamic data, such as optical flow, gives extra information to the network and an understanding of the physical world that the network does not have with only the rgb video stream. The OF, which represents the movement through the displacement of the pixels from one image to another, is encoded in the Cartesian coordinate system $V=(v_x,v_y)^T$ with $v_x$ the horizontal displacement and $v_y$ the vertical displacement. The optical flow values are then normalized between -1 and 1.

In~\cite{PeICIP}, several optical flow estimation methods are compared using Mean Squared Error (MSE) of motion compensation, angular error (AE) and end-point error (EPE) quality metrics on two datasets:
\begin{itemize}
    \item Sintel Benchmark ~\cite{Dataset:LargeOpticalFlowDataset}, dataset of synthetic videos with available reference optical flows (comapred with MSE, AE and EPE),
    \item and \texttt{TTStroke-21}~\cite{PeMTAP}, recorded in natural conditions with strong flickering due to synthetic light (compared with MSE only since reference flow is not available).
\end{itemize}
In the same work, different normalization methods are also tested for classification. According to their results, both of these modalities, optical flow estimator and normalization method, are of primary importance for classification. Indeed, the accuracy of the classification on the test sets of \texttt{TTStroke-21} varies, according to the best performances for each optical flow estimator, from $41.4$\% for DeepFlow method~\cite{OF:DeepFlow} to $74.1$\% for Beyond Pixel method~\cite{OF:BP} (BP). Even if BP is sensitive to flickering, it is able to capture fine details, such as the motion of the ball, contrarily to DeepFlow method, explaining such gap between performances of those two estimators. On the other hand, the normalization methods allow fro boosting performances from $44$\%  with ``Max'' normalization method to $74.1$\% with ``Normal'' normalization method. The ``Max'' normalization method strongly reduces the magnitude of most of motion vectors and therefore increases also the inter-similarity of the strokes; while the ``Normal'' normalization method increases the magnitude of most vectors and leaves room for inter-dissimilarity.

In the light of these results, we use the Beyond Pixel method~\cite{OF:BP}, based on iterative re-weighted least square solver, to estimate the optical flow and normalize it using the ``Normal'' normalization method as described in eq. \ref{chap10:eq:Normal}. 

\begin{equation}
\label{chap10:eq:Normal}
	\begin{array}{l}
    	v' = \frac{v}{\mu + 3\times \sigma}\\
    	v^N(i,j) = \left\{
    	\begin{array}{ll}
    	v'(i,j) & \mbox{if } |v'(i,j)| < 1 \\
    	SIGN(v'(i,j)) & \mbox{otherwise.}
    	\end{array}
    	\right.
	\end{array}
\end{equation}
 where $v$ and $v^N$ represent respectively one component of the optical flow $\textbf{V}$ and its normalization, $\mu$ and $\sigma$ are mean value and standard deviation of the component.
 
The optical flow is then filtered by considering only the optical flow of the foreground  using the method of Zivkovic and Van der Heijden~\cite{Zivkovic}.

\subsubsection*{3D Attention - what could it bring?} 

Attention mechanisms, in classification problem from rgb images, are used to determine which part of information is useful and/or needed to classify an image. Such attention can be obtained by recording the gaze fixations of individuals when looking at the image and performing the same classification task~\cite{Attention::saliency} as a CNN classifier, to create saliency map on image and then propagate it through the layers of the CNN. 
In DNNs, internal attention mechanisms have become popular. We distinguish two of them: i) global attention which expresses the contribution of feature channels along convolutional layers into decision making for image classification task~\cite{Attention::squeeze} and ii) local attention, which focuses on important features in the channels. 
\par
When importance of feature channels has to be computed, the processing consists of three steps: i) \textit{squeeze} (synthesis), ii) \textit{excitation} and iii)  \textit{feature scaling}. Thus a small network of neurons learns a weighting coefficient for each channel at each layer and outputs the characteristic channels thus weighted to the next layer. 
\par
For the local attention, the authors of~\cite{Attention::RANImageClassification} use the principles of residual neural networks to propose ``residual'' learning of the attention masks incorporated in the convolution layers in both forward and backward propagation, which leads to better robustness to noise. By minimizing the objective function by gradient descent, the attention mechanisms are implicitly introduced via the derivative calculation where the weighted characteristics are used. 
``Teacher-student'' networks is another way to introduce attention in the layers of a classification network. In~\cite{Attention::Transfer} the ``Teacher'' network is the one that learns attention and guides the "student" network for the image classification task. Such works which focus on image classification inspire the design of spatio-temporal attention to tackle the action classification problem.
\par
\begin{figure}[ht]
		\centering
		\includegraphics[width=\linewidth]{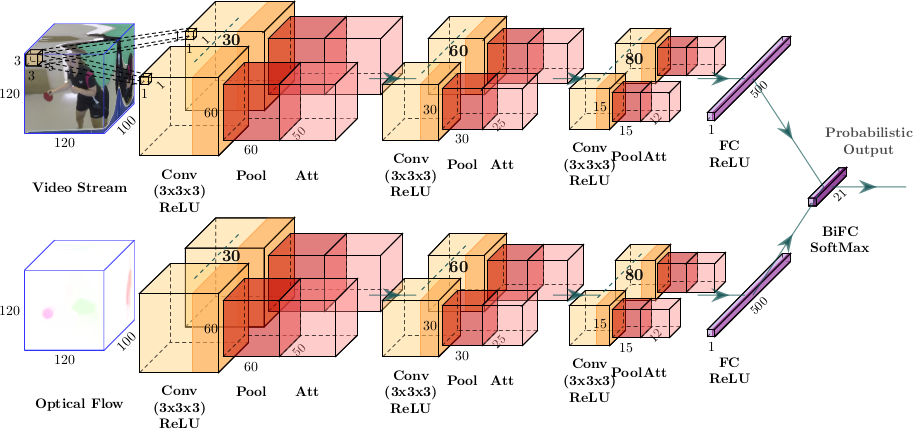}
		\caption{3D attention block architecture}
		\label{chap10:fig:TSTCNN_Attention}
\end{figure}

In~\cite{Attention::Pyramid}, the authors use spatial attention mechanism based on feature pyramids and construct the temporal attention by aggregation of the attention maps obtained spatially, in the temporal domain. Then, based on their work, the authors of~\cite{Attention::Res3ATN} build 3D convolutional blocks and incorporate them in a 3DResNet network for 3D gesture recognition. However, as in~\cite{Attention::Pyramid}, the authors do not use the motion information explicitly. In~\cite{PeICPR2020}, we introduce the attention blocks through our TSTCNN into the two video streams: the branch containing the spatial information (RGB) and the branch containing the temporal information (optical flow) as depicted in fig.~\ref{chap10:fig:TSTCNN_Attention}. The optical flow plays the discriminating role for classification and does not act only as localization information.

\begin{figure}[htb]
		\centering
		\includegraphics[width=\linewidth]{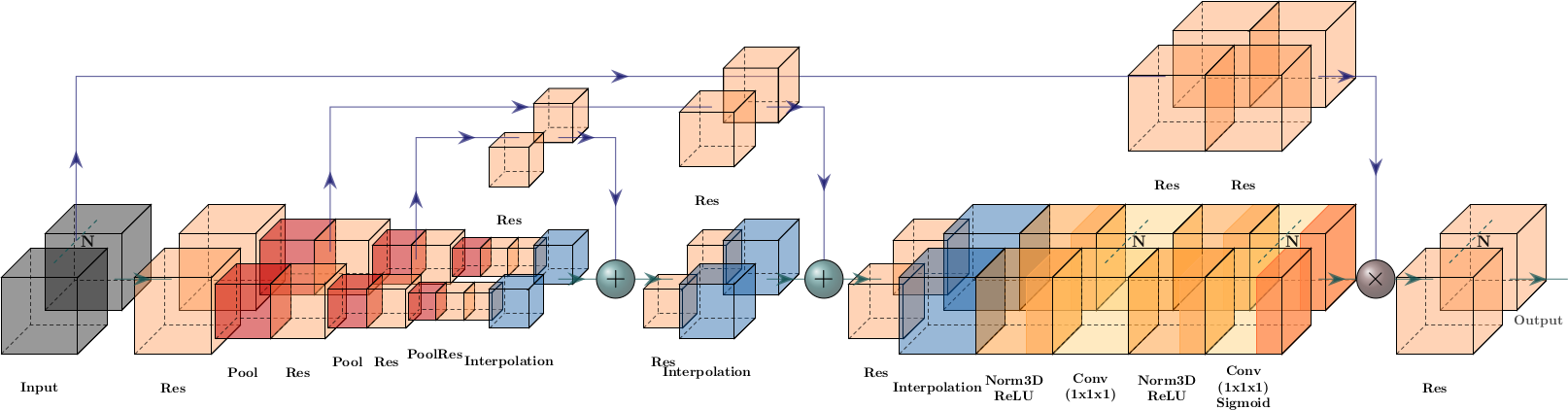}
		\caption{TSTCNN with 3D attention blocks after each Max Pool layer}
		\label{chap10:fig:Attention}
\end{figure}

\begin{figure}[ht]
		\centering
		\includegraphics[width=0.7\linewidth]{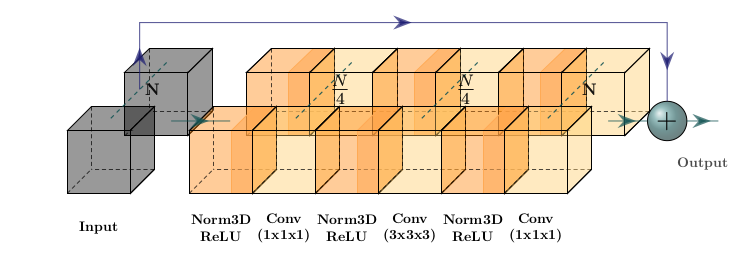}
		\caption{3D residual block architecture}
		\label{chap10:fig:ResidualBlock}
\end{figure}

As depicted in fig. \ref{chap10:fig:Attention}, the attention mechanism uses several 3D residual blocks, illustrated in fig. \ref{chap10:fig:ResidualBlock}.

The implemented residual block is inspired from the work carried out in 2D in~\cite{NN:ResNet} and has been extended and adapted in 3D with a 4D data block of size ($N \times W \times H \times T$) representing respectively the number of channels, the two spatial dimensions and the temporal dimension. Input data are then processed by 3 successive layers $f_{conv_i} , i=1,...,3 $, see eq.~\ref{chap10:eq:floatingbranch3}, with respectively $\frac{N}{4}$,  $\frac{N}{4}$ and $N$ filters of size ($1\times1\times1$), ($3\times3\times3$) and ($1\times1\times1$). Their output is then summed with the input data to build the output of the 3D residual block.

\begin{equation}
\label{chap10:eq:res}
    res(x) = f_{conv_3}(f_{conv_2}(f_{conv_1}(x))) +  x
\end{equation}

The 3D batch normalization is performed channel by channel over the batch of data. If we have $x = (x_{1}, x_{2}, ... , x_{N_{channels}})$, then the normalization is $F_n(x) = (f_n(x_{1}), f_n(x_{2}), ... , f_n(x_{N_{channels}}))$ with:

\begin{equation}
\label{chap10:eq:batchnorm3d}
    f_n(x_i) = \frac{x_i - \mu_i}{\sqrt{\sigma_i^2 \oplus \epsilon}} * \gamma_i \oplus \beta_i
\end{equation}

with $i = 1, ... N_{channels}$, $\mu_i$ and $\sigma_i$ the mean and standard deviation vectors of $x_i$ computed over the batch,$\oplus$ is addition of a scalar to each vector coordinate,  $\gamma_i$ and $\beta_i$ are learnable parameters per channel and the division by $\sqrt{\sigma_i^2\oplus\epsilon}$ is element-wise. Here, $N_{channels}=N$ or $\frac{N}{4}$, depending on the normalization position in the residual block.

The 3D attention blocks illustrated in Figure~\ref{chap10:fig:Attention} are inspired from the work carried out in 2D in~\cite{Attention::RANImageClassification}.

A 3D attention block takes as input a 4D data block of size ($N \times W \times H \times T$). In this block, all convolutions are performed with the same number of filters, $N$, to maintain the dimension of the processed data. The input data are processed by the first 3D residual block, denoted as ``\emph{res}''. The network then splits in two branches: the trunk branch consisting of 2 successive 3D residual blocks, described by equation~\ref{chap10:eq:trunkbranch}; and the soft floating mask branch, see lowest position in Figure~\ref{chap10:fig:Attention}), described through the equations~\ref{chap10:eq:floatingbranch1}, \ref{chap10:eq:floatingbranch2}, \ref{chap10:eq:floatingbranch3}. Its role is to accentuate the features generated by the trunk branch. Those two branches are merged as described in equation~\ref{chap10:eq:attbranches}.

\begin{equation}
\label{chap10:eq:trunkbranch}
    branch_{trunk}(.) = res(res(.))
\end{equation}

The soft mask branch is constituted of several 3D residual blocks followed by Max Pooling layers, denoted as ``\emph{MaxP}''. It increases the reception field of convolutions using a bottom-up architecture, denoted as $f_{bu}(.)=res(MaxP(.))$. The lowest resolution is obtained after $3$ Max Pooling steps.

\begin{equation}
\label{chap10:eq:floatingbranch1}
\begin{array}{l}
    x_1 = f_{bu}(res(Input))\\
    x_2 = f_{bu}(x_1))\\
    x_3 = f_{bu}(x_2))
\end{array}
\end{equation}

The information is then extended by a symmetrical top-down architecture, $f_{td}(.) = Inter(res(.))$, to project the input features of each resolution level. ``\emph{Inter}'' denotes the trilinear interpolations used for up-sampling. Two skipped connections are used for collecting information at different scales.

\begin{equation}
\label{chap10:eq:floatingbranch2}
\begin{array}{l}
   y_1 = f_{td}(x_3) + res(x_2)\\
    y_2 = f_{td}(y_1) + res(x_1)\\
    y_3 = f_{td}(y_2)
\end{array}
\end{equation}

The soft mask branch is then composed of 2 successive layers. Each includes a 3D batch normalization, denoted as $F_n(.)$ as described by equation~\ref{chap10:eq:batchnorm3d}, followed by a ReLU activation function and a convolution layer with kernel sizes ($1\times1\times1$). This is expressed by equation~\ref{chap10:eq:fconv}:

\begin{equation}
\label{chap10:eq:fconv}
    f_{conv}(.) = conv(ReLU(F_n(.)))
\end{equation}

It ends with a sigmoid function, denoted as ``Sig'', to scale values between $0$ and $1$. These two layers are depicted on the right of the lowest branch in figure~\ref{chap10:fig:Attention} and are expressed by equation~\ref{chap10:eq:floatingbranch3}. 

\begin{equation}
\label{chap10:eq:floatingbranch3}
    branch_{fmask}(Input) = Sig(f_{conv}(f_{conv}(y_3)))
\end{equation}

The output of the trunk branch is then multiplied term by term by $(1 \oplus  branch_{fmask} (Input))$ where $branch_{fmask}(Input)$ is the output of the mask branch. The result is then processed by the last 3D residual block $res(.)$ which ends the attention block, see equation~\ref{chap10:eq:attbranches}. 

\begin{equation}
\label{chap10:eq:attbranches}
    y = res(branch_{trunk}(Input) \odot (1 \oplus branch_{fmask}(Input)))
\end{equation}
Here the $\odot$ is an element-wise multiplication and $\oplus$ is an addition of a scalar to each vector component as defined above.

\subsection{Results} 

In Table~\ref{Chap8:table:acc_all}, we compare the models in terms of accuracy for the pure classification task. In order to have an overall view, comparison is done with the models using three attention blocks (one after each max pooling layer) with the models and  without them.

\begin{table}[htbp]
	\centering
	\caption{Comparison of the classification performances for models using attention mechanism after convergence in terms of accuracy.}
	\label{Chap8:table:acc_all}
	\begin{tabular}{ccccc}
		\cline{3-5}\noalign{\smallskip}
		& & \multicolumn{3}{c}{\textbf{Accuracies in \%}}\\
		\noalign{\smallskip}\hline\noalign{\smallskip}
		\textbf{Models} &\textbf{Epochs} &\textbf{Train} &\textbf{Val}   &\textbf{Test}\\
		\noalign{\smallskip}\hline\noalign{\smallskip}
    	
    	RGB-I3D         &$778$  &$98.3$   &$72.6$     &$69.8$      \\
    	RGB-STCNN       &$1665$ &$96.7$   &$88.7$   &$89.8$	    \\
    	\bf RGB-STCNN with Attention    &$524$  &$96.5$    &$88.3$    &$\bf93.2$ \\[5pt]
    	Flow-I3D                   &$1112$ &$98.8$     &$74.8$     &$73.3$ \\
        Flow-STCNN  	            &$1449$ &$97.5$     &$79.6$     &$75.9$    \\
        \bf Flow-STCNN with attention   &$732$  &$96.4$	    &$83.5$     &$\bf90.7$  \\[5pt]
    	Two-Stream I3D             & -     &$99.2$     &$76.2$     &$75.9$  \\
        LF-STCNN                  & -     &$97$       &$88.7$     &$89.8$      \\
        
        LF-STCNN with attention   &-     &$97$      &$88.7$    &$94.9$   \\
        T-STCNN                   &$1784$ &$95.8$     &$87.8$     &$93.2$    \\
        
        \bf T-STCNN with attention                  &$591$         &$97.3$   &$87.8$  &$\bf95.8$ \\
        \noalign{\smallskip}\hline\noalign{\smallskip}
    \end{tabular}
\end{table}

The presented models using spatio-temporal convolutions shallower than the I3D models prove to be more efficient to classify fine-grained actions on a challenging dataset with limited amount of samples. The fine-grained aspect of the task seems to be dealt better if the model is not too deep. Indeed, the deepness of the I3D models is efficient on task were a great number of objects and scenes need to be recognized in order to classify coarse-grained actions, but it seems to be less effective for the fine-grained task.

\par

Furthermore, the amount of samples and the fact that the models are not pre-trained, makes the task even harder. The overfitting problem is more noticeable on the very deep models I3D. However, the STCNN models using attention mechanisms do not suffer from this: the performances are improved and the convergence is faster. Best performances are observed for the Twin model using attention mechanism. The intermediate fusion allows a better combination of each branches in order to perform classification. A late fusion approach seems efficient but is 1\% behind.

\section{Conclusion and Perspectives} 

In this chapter, we have shown that the features developed to perform for action classification in videos have evolved from handcrafted to deep learnt with the increasing complexity of datasets and their increasing number of video samples. Furthermore, the increasing capacity of CPUs and GPUs to process large amount of data have allowed deep learning methods to move from 2D to 3D convolutions. 3D convolutions for action classification from videos have proven to be efficient tools to capture efficiently spatio-temporal features in order to perform classification.
\par
The incorporation of attention mechanisms, through attention blocks, helps in the process by highlighting the discriminant features, boosting the convergence speed and classification performances. Such behaviour was observed on the fine-grained dataset TTStroke-21. It has also been noted that the deepness of the implemented models needs to be adapted to the classification task and the dataset in order to avoid overfitting.
\par


\bibliographystyle{alpha}
\bibliography{biblio}
\end{document}